\pdfoutput=1
\documentclass[11pt]{article}

\usepackage{EACL2023}
\usepackage{booktabs}
\usepackage{amsmath}
\usepackage{times}
\usepackage{float}
\usepackage{latexsym}
\usepackage{graphicx}
\usepackage[T1]{fontenc}
\usepackage[utf8]{inputenc}

\usepackage{microtype}

\usepackage{inconsolata}

\title{Detecting Structured Language Alternations in Historical Documents by Combining Language Identification with Fourier Analysis}

\author{Hale Sirin \\
  Center for Digital Humanities \\
  Johns Hopkins University \\
  \texttt{hsirin1@jhu.edu} \\\And
  Sabrina Li \\
  Center for Digital Humanities \\
  Johns Hopkins University \\
  \texttt{sli159@jhu.edu} \\\And
  Tom Lippincott \\
  Center for Digital Humanities \\
  Johns Hopkins University \\
  \texttt{tom@cs.jhu.edu} \\}

\begin{document}
\maketitle
\begin{abstract}
In this study, we present a generalizable workflow to identify documents in a historic language with a nonstandard language and script combination, Armeno-Turkish. We introduce the task of detecting distinct patterns of multilinguality based on the frequency of structured language alternations within a document. 
\end{abstract}

\section{Introduction }

This work emerges from the goal to create a corpus in Armeno-Turkish—vernacular Turkish written in Armenian script. This historic language was actively used from the early 18th century to the early 20th century in a variety of locations in the Middle East, Europe and the US, including Istanbul, Venice, Vienna and Boston \cite{ATdev}.
There are lists of works in Armeno-Turkish available \cite{stepanyan}, but searching HathiTrust manually points to the existence of works that are not recorded in these lists due to challenges in the bibliographical recording of these works in library catalogues (missing titles in the original script, wrong or missing language labels). Unable to collate works in Armeno-Turkish by using the metadata, we used language identification. This process did not produce a clean dataset in Armeno-Turkish, but showed that a significant portion of these works are multilingual. Based on this observation, this study is a first attempt at modeling some of the interesting multilingual phenomena with structured language alternations that emerge in this process: bilingual translations, dictionaries, original-language text followed by commentary in a different language, language study books. 

As opposed to unstructured code switching (oral interviews), structured multilingual patterns involve an organized alternation of two or more languages. These language alternations may occur at different frequencies (every sentence, paragraph, every page, every chapter). A structured language alternation may serve various purposes, including making the content available in multiple languages in a legal document to reach its target audiences, as shown in the two-column Ottoman legal code in Armenian and Armeno-Turkish in Figure 1. 
\begin{figure}
\centering
\includegraphics[width=.32\textwidth]{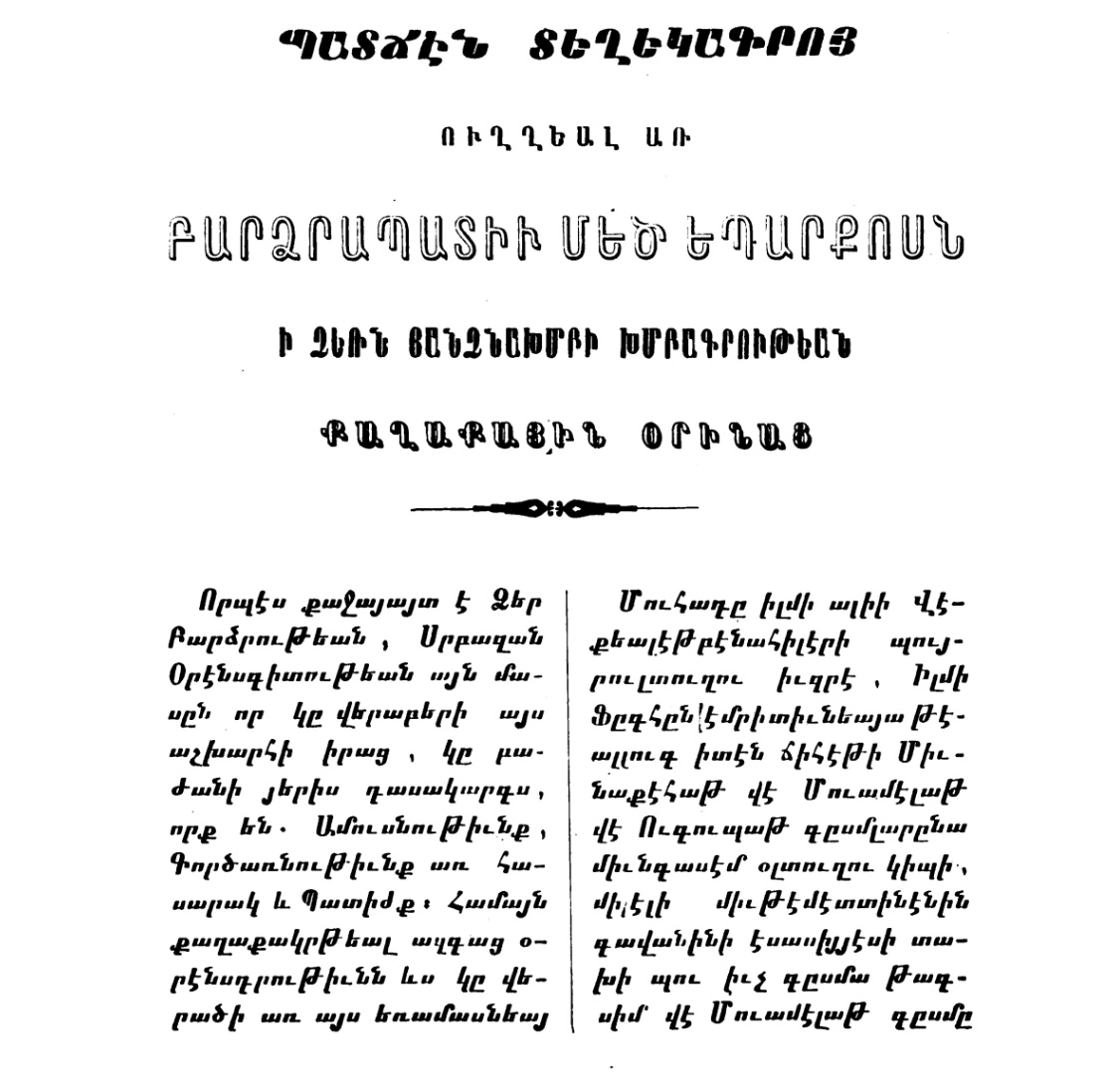}
\caption{The first page of the Ottoman legal code, \textit{Mejelle}, published in 1889 in a bi-column bilingual format, Armenian on the left and Armeno-Turkish on the right \cite{mejelle}.}
\label{fig:words}
\end{figure}

Especially for historic languages, detecting structured language alternations is valuable for identifying clean monolingual segments which can then be used to create new resources for NLP tasks. This analysis also provides insight into material history of physical books and translation studies, by showcasing different formats of page segmentation in structured multilingual books (bi-column, top-bottom) \citep{segment, sarah}. 
This project introduces the task of detecting structured language alternations and makes the following contributions:  

\begin{itemize}

    \item We introduce an experiment that maps the language alternations in the time domain to the frequency domain to detect different patterns of structured language alternations in a corpus, and show that unsupervised clustering applied to the frequency spectra can be a simple and efficient first step in grouping documents with different patterns of structured language alternations. 
    
    \item We present a more comprehensive and nuanced Armeno-Turkish corpus from the HathiTrust Digital Repository, and compare the performance of a character n-gram model and a trained neural model for language identification of a historic language with a nonstandard language and script combination. 
    
\end{itemize}

\section{Background}
\subsection{Language ID}

Language identification is the task of determining the language(s) of a document and a crucial step in document classification in historical research. Character-based n-gram models are a performant statistical approach \cite{Cavnar1994NgrambasedTC}, and recent neural models offer a fast and efficient solution \cite{joulin2016fasttext}. 

While language identification of a document is mostly regarded as a solved task \cite{McNamee}, a near-perfect performance is only achieved when certain assumptions are made regarding the quantity and the quality of the data, and the monolinguality of the documents. However, historic and multilingual documents in low-resource languages motivate different approaches to language identification \cite{langIDsurvey}.  Multilinguality of these documents is frequently overlooked, even though historic languages in non-standard scripts, such as Armeno-Turkish, represent territories that were predominantly multilingual. Despite the perceived monolingualism of 18th-and-19th-century books that are published in “national print-languages” \cite{anderson2006imagined}, multilingual activity persisted and even flourished during this period in commercial, legal, cultural and literary domains \cite{+2023}. 

Research in multilingual language identification and code switching generally focuses on identifying the language, but not the relative location of these languages in each document \cite{lui-etal-2014-automatic}. \citet{kevers-2022-coswid} locates code switches, but primarily in multilingual documents when language diversity is unstructured. In this study, we focus on distinct patterns of structured language alternations that emerge in historical datasets (religious commentary, language study books, bilingual legal documents) in Armeno-Turkish, a low-resource language that falls outside the “national-print language” category.
\subsection{Frequency Analysis}
Discrete Fourier Transform (DFT) converts a time-domain sequence to a frequency domain sequence. It's defined by equation 1, mapping a sequence of N numbers $x_0,x_1 \ldots, x_{N-1}$ to a new sequence of N numbers, for $0 \leq k \leq N - 1$.

\begin{equation}   \label{Equation 1}
    X_k =   \sum_{n=0}^{N-1} x_n e^{-2\pi ikn/N }
\end{equation}

Fast Fourier Transform (FFT) is an efficient algorithm used to calculate the DFT, reducing the time complexity from $O(N^2)$ to $O(N $ {log} $ N)$ \cite{ct1965}.
Fourier transform has a wide range of applications in NLP. In this study, we approached the probability of a language label as a discrete signal at 50-word time steps in a document. Figure 2 shows a simulation of this approach, representing an idealized alternation of one language and another language as an array of alternating 0s and 1s and plotting its frequency domain representation using the Fourier Transform.

\begin{figure}[H]
\includegraphics[width=.45\textwidth]{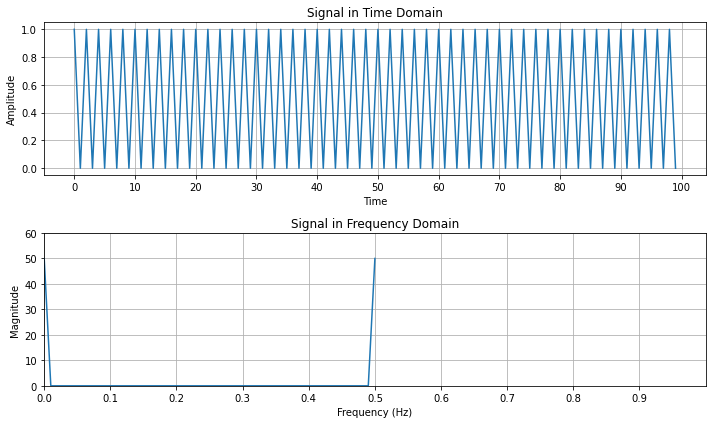}
\caption{Time domain and frequency domain representations of an alternating discrete signal.}
\label{fig:words}
\end{figure}

\section{Materials and Methods \footnote{Code for the full experiment is available at \url{https://github.com/comp-int-hum/Armeno-Turkish-Collection}}}

\subsection{Data}

HathiTrust Digital Library (HT) \cite{htc} offers unprecedented access to scholars who work with text as data in a variety of disciplines. However, corpus construction is a significant challenge when working with historic languages and multilingual documents (due to missing language and script information, OCR errors) and the overarching language label, even when it is correct, does not provide information regarding the multilingual composition of a given document. The MARC 21 Bibliographic Record offers a limited structure to include script-related information, but it is not put to use systematically by cataloging librarians. The script information for the Armeno-Turkish works, if there is any, is occasionally found in the Notes section of the MARC record. This means that researchers' only recourse is to browse works in Armenian and in Turkish manually with the hopes of encountering Armeno-Turkish works.  
\subsection{Language ID Experimental Setup}
We start by creating a training dataset of works labeled according to both language and script from the HT. For Armeno-Turkish, we use 97 expert-labeled documents. For negative examples, we first use the HT's MARC index to create a reverse mapping of languages (as assigned by librarians at the contributing institution) to documents, skipping anything dated before 1500 CE.  We remove languages with less than 100 documents or whose code is not valid ISO-639. To ensure diverse temporal representation of each language, we split the range from the earliest to latest document in that language into 5 buckets covering equal time periods, and randomly select one document from each bucket.  We then select an additional 5 documents at random from the overall remaining set, for a total of 10 documents per language. We split each document into sub-documents of contiguous script according to the Unicode script specification. This process results in 118 unique script and language combinations, serving as our training data. For the positive examples we only keep the Armenian script, since these are hand-annotated and does not rely on MARC metadata. At test time, we segment the document into smaller sections. By recording the segmentation offsets, the original documents can be reconstructed with the inferred language information. 

We compare a trigram character language model with FastText neural language ID model \cite{joulin2016fasttext}, trained on our labeled dataset in Table 1. We create a dataset of all documents in the HTC tagged as Turkish (tur), Ottoman Turkish (ota), or Armenian (arm), a total of 18367 records. 

We apply the trained FastText model with a 0.91 fscore on the test set, to all documents in the HTC tagged as Turkish (tur), Ottoman Turkish (ota), or Armenian (arm) in 50-word windows. This process yields 95 works with Armeno-Turkish as the majority language label.  

\begin{table}[t]
\small
\centering
\begin{tabular}{|c|c|}
\hline
Model & Fscore (macro) \\
\hline
Multinomial NB   & 0.67\\
Char N-gram  & 0.83 \\
Trained FastText & 0.91 \\
\hline
\end{tabular}
\caption{Lang ID Model Performance Comparison}
\end{table}

\subsection{Frequency Analysis}
The output of the FastText model is a grid of probability distributions over all possible language\_script labels for each 50-word window in a document. In order to reach our goal of bringing out the periodic phenomena in these 95 works, we simplify this fairly noisy information. Since it is unclear how well-calibrated the model is, we calculate the majority language label of the whole document, and use the probability value of that language for each chunk. This  language-agnostic approach radically simplifies the initial big grid of the full probability distribution, into a sequence of probabilities of that majority language for each 50-word chunk.  

We transform each probability distribution array into a frequency domain, using FFT, and cluster each frequency spectrum using k-means clustering. Frequency domain transformation allows us to compare signals of different lengths. 

\section{Results and Discussion}
The clustering of the frequency spectra yielded three coherent groups: 
\begin{enumerate}
    \item Works that are predominantly in Armeno-Turkish (59 documents) 
    \item Works that are bilingual, alternating between Armeno-Turkish and another language (10 documents: Language textbooks: alternating every few sentences, Bilingual editions: alternating every column or every page) 
    \item  Works that are multilingual in languages other than Armeno-Turkish (13 documents: Language textbooks in Armenian-Greek, Armenian-Russian, Armenian-French)
\end{enumerate}

\begin{figure}[H]
\centering
\includegraphics[width=.35\textwidth]{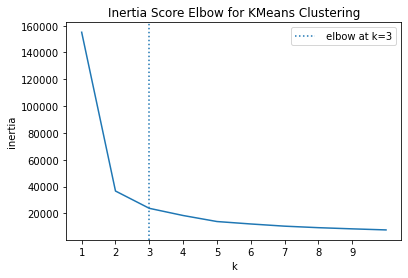}
\caption{Visualization of k-elbow inertia metric for
optimal k in k-means clustering.}
\label{fig:words}
\end{figure}
As visualized in Figure 3, we selected the number of clusters based on the inertia metric for optimal k. While the clusters are relatively coarse-grained, this is a fast and efficient approach to historical datasets with unreliable metadata and high variation in genre and language composition. The frequency analysis combined with segmented language identification is a promising venue to explore documents in historic languages, since it lets us divide the corpora automatically into more distinct categories, revealing a variety of genres. Preserving the diversity of genres is valuable for low resource situations in which there is a risk of certain genres dominating the fine-tuning material.

This experiment identified 30 new records in Armeno-Turkish that were not in the training set, including translations of the Bible, dictionaries, textbooks for learning foreign languages, and legal documents. 

\subsection{Error Analysis}

\begin{figure}[H]
\centering
\includegraphics[width=.35\textwidth]{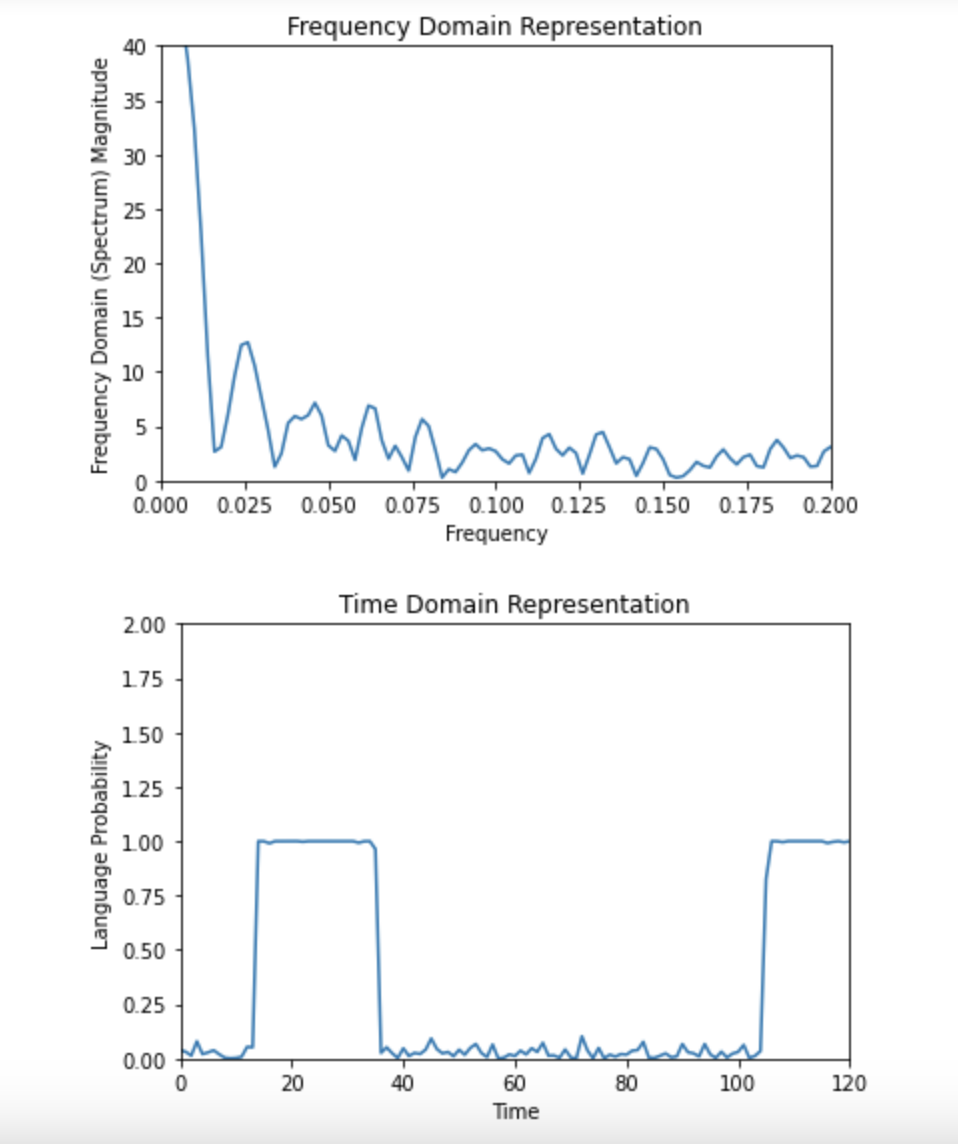}
\caption{Time domain and frequency domain representations of the alternating language probability signal in a section of the monolingual book with page segmentation shown in Figure 4.}
\label{fig:words}
\end{figure}

For example, Figure 4 shows a book that is entirely in Armeno-Turkish, but is segmented into four parts, with the lower two segments in a smaller font and occasionally in a different typeface, resulting in a significantly worse OCR output periodically. This creates a falsely identified language alternation pattern. Figure 5 shows the frequency and time domain representations of the Armeno-Turkish probability in the same book. In comparison, figure 6 shows the time-domain and frequency-domain signal representations of an actual bilingual book with an alternation pattern every sentence. 

\begin{figure}[H]
\centering
\includegraphics[width=.37\textwidth]{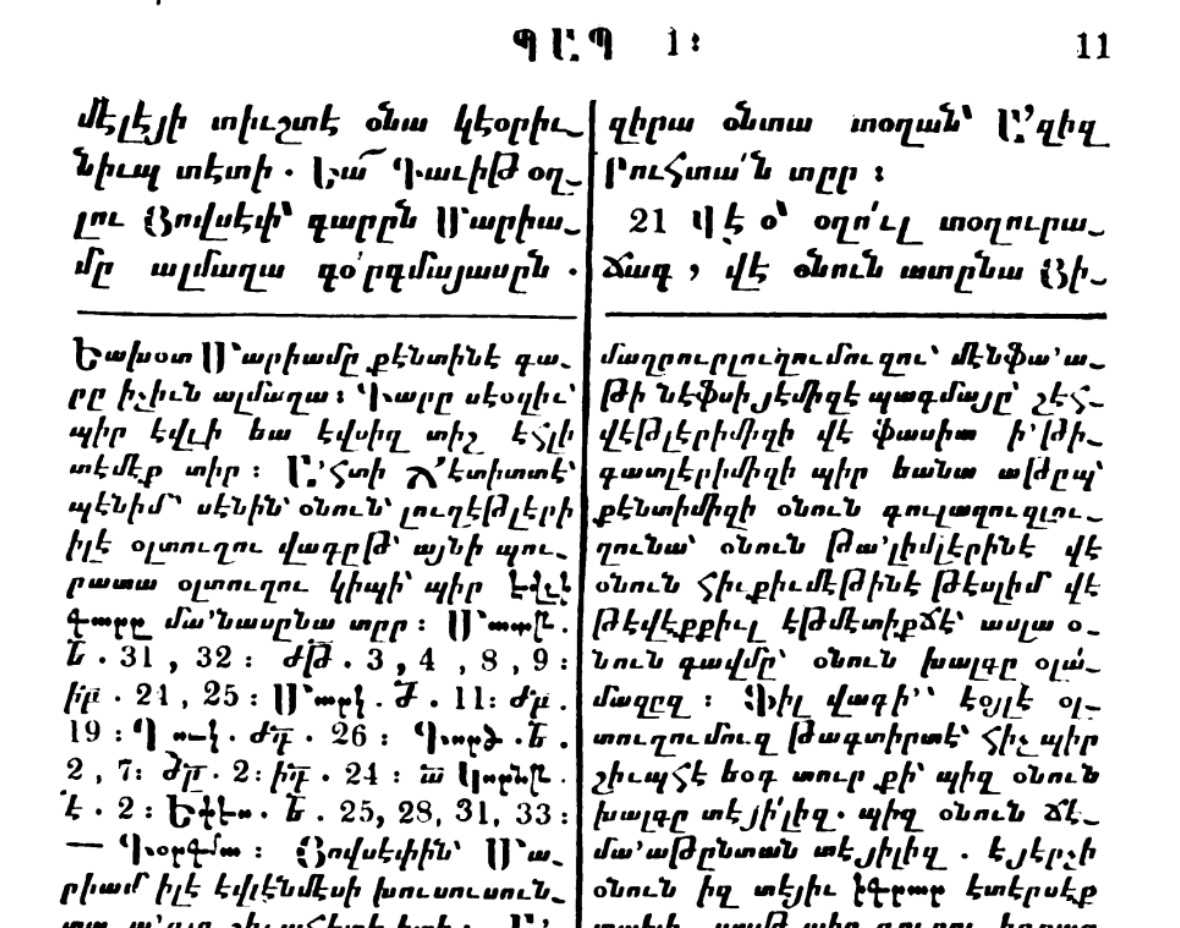}
\caption{Page segmentation in the book, \textit{Commentary On the Gospel of Matthew}, in Armeno-Turkish. \cite{segmentedgospel}}
\label{fig:words}
\end{figure}

\begin{figure}[H]
\centering
\includegraphics[width=.35\textwidth]{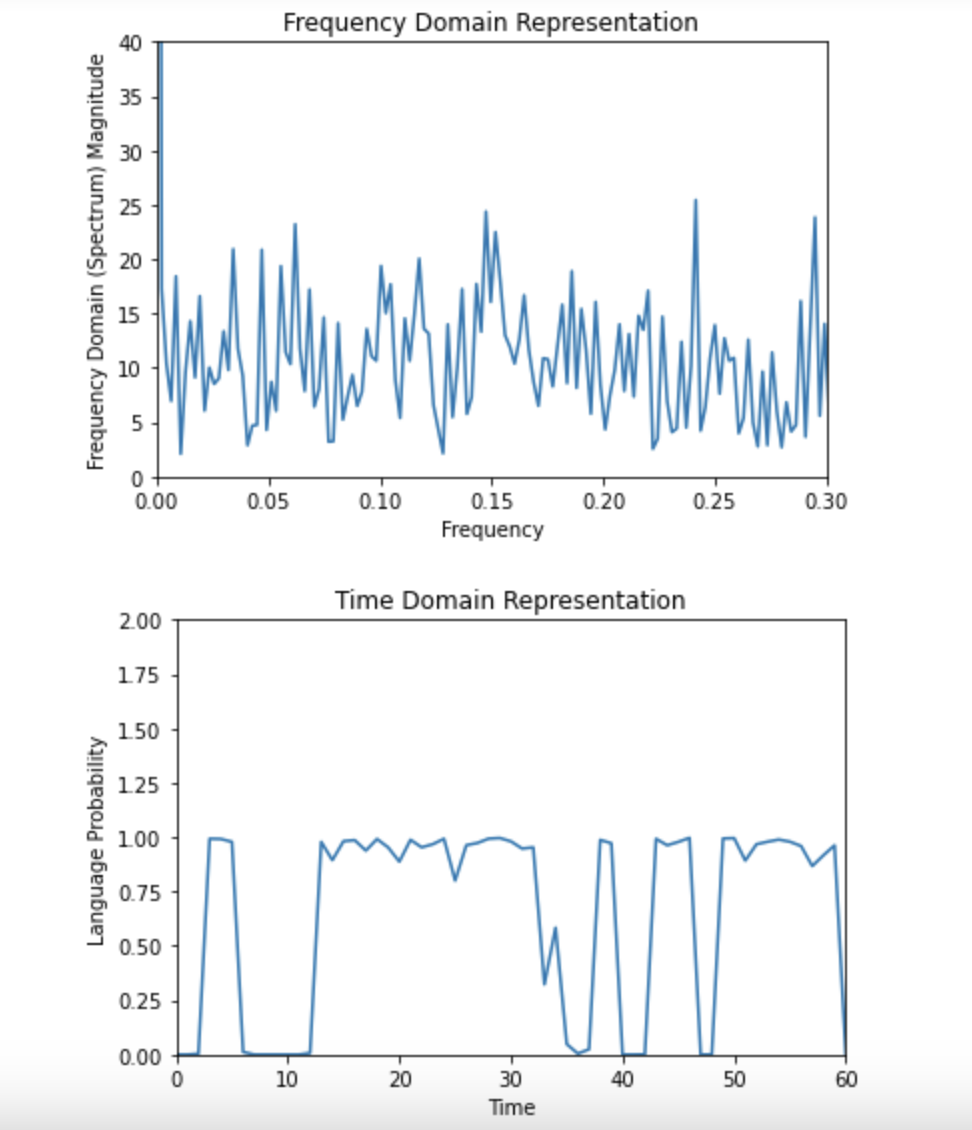}
\caption{Time domain and frequency domain representations of the alternating language probability signal in a section of a bilingual book that alternates between Armenian and Armeno-Turkish every sentence. \cite{Erger}}
\label{fig:words}
\end{figure}

The patterns of language alternation that emerge are not very fine-grained, due to the high degree of noise in the HathiTrust corpus. In some cases, this leads to alternations in the lang ID probability results, not due to a change in the language, but due to a periodic noise in the post-OCR text (footnotes in smaller font, highly segmented pages). 

\section{Future Work}
We plan to expand the frequency analysis approach by using a cleaner dataset and including different languages with the goal of reaching a more nuanced classification at the word or sentence level (such as dictionaries).  This process has a potential to be applied as a feature extractor for a downstream classification task. Training a classifier on clean data, where the patterns of structured language alternations are known, could lead to a specific classifier (dictionary, bilingual, annotated edition). This process would require clean data, but we hypothesize that a model trained on carefully annotated high-resource data could then be used on a low-resource language, since the periodic signal would be the same across languages. 

\bibliographystyle{acl_natbib}
\bibliography{latech_paper.bib}

\begin{thebibliography}{17}
\expandafter\ifx\csname natexlab\endcsname\relax\def\natexlab#1{#1}\fi

\bibitem[{Erg(1881)}]{Erger}
 1881.
\newblock \emph{Erger Tghayots' Hamar}.
\newblock Konstandnupolis: Tpagrut'iwn Aramean.

\bibitem[{mej(1889)}]{mejelle}
 1889.
\newblock \href {http://hdl.handle.net/2027/nyp.33433021340926} {\emph{Mejelle [Ottoman Civil Law.]}}.

\bibitem[{Anderson(2006)}]{anderson2006imagined}
B.~Anderson. 2006.
\newblock \href {https://books.google.com/books?id=nQ9jXXJV-vgC} {\emph{Imagined Communities: Reflections on the Origin and Spread of Nationalism}}.
\newblock ACLS Humanities E-Book. Verso.

\bibitem[{Cavnar and Trenkle(1994)}]{Cavnar1994NgrambasedTC}
William~B. Cavnar and John~M. Trenkle. 1994.
\newblock \href {https://api.semanticscholar.org/CorpusID:170740} {N-gram-based text categorization}.

\bibitem[{Cooley and Tukey(1965)}]{ct1965}
James~W. Cooley and John~W. Tukey. 1965.
\newblock \href {https://www.ams.org/journals/mcom/1965-19-090/S0025-5718-1965-0178586-1/S0025-5718-1965-0178586-1.pdf} {An algorithm for the machine calculation of complex {F}ourier series}.
\newblock \emph{Mathematics of Computation}, 19(90):297--301.

\bibitem[{Der~Matossian(2020)}]{ATdev}
Bedross Der~Matossian. 2020.
\newblock The development of armeno-turkish (hayatar t‘rk‘erēn) in the 19th century ottoman empire: Marking and crossing ethnoreligious boundaries.
\newblock \emph{Intellectual History of the Islamicate World}, 8(1):67--100.

\bibitem[{Goodell(1851)}]{segmentedgospel}
William Goodell. 1851.
\newblock \emph{Madtéos Injilinin Tefsiri: [Commentary on the Gospel of Matthew]}.
\newblock Smyrna: William Griffith.

\bibitem[{HathiTrust Foundation()}]{htc}
HathiTrust Foundation. 2023.
\newblock \href {https://www.hathitrust.org} {{HathiTrust Digital Library}}.

\bibitem[{Jauhiainen et~al.(2018)Jauhiainen, Lui, Zampieri, Baldwin, and Lindén}]{langIDsurvey}
Tommi Jauhiainen, Marco Lui, Marcos Zampieri, Timothy Baldwin, and Krister Lindén. 2018.
\newblock \href {http://arxiv.org/abs/1804.08186} {Automatic language identification in texts: A survey}.

\bibitem[{Joulin et~al.(2016)Joulin, Grave, Bojanowski, Douze, J{\'e}gou, and Mikolov}]{joulin2016fasttext}
Armand Joulin, Edouard Grave, Piotr Bojanowski, Matthijs Douze, H{\'e}rve J{\'e}gou, and Tomas Mikolov. 2016.
\newblock Fasttext.zip: Compressing text classification models.
\newblock \emph{arXiv preprint arXiv:1612.03651}.

\bibitem[{Kevers(2022)}]{kevers-2022-coswid}
Laurent Kevers. 2022.
\newblock \href {https://aclanthology.org/2022.sigul-1.15} {{C}o{S}w{ID}, a code switching identification method suitable for under-resourced languages}.
\newblock In \emph{Proceedings of the 1st Annual Meeting of the ELRA/ISCA Special Interest Group on Under-Resourced Languages}, pages 112--121, Marseille, France. European Language Resources Association.

\bibitem[{Lui et~al.(2014)Lui, Lau, and Baldwin}]{lui-etal-2014-automatic}
Marco Lui, Jey~Han Lau, and Timothy Baldwin. 2014.
\newblock \href {https://doi.org/10.1162/tacl_a_00163} {Automatic detection and language identification of multilingual documents}.
\newblock \emph{Transactions of the Association for Computational Linguistics}, 2:27--40.

\bibitem[{McConnaughey et~al.(2017)McConnaughey, Dai, and Bamman}]{segment}
Lara McConnaughey, Jennifer Dai, and David Bamman. 2017.
\newblock \href {https://doi.org/10.18653/v1/D17-1077} {The labeled segmentation of printed books}.
\newblock In \emph{Proceedings of the 2017 Conference on Empirical Methods in Natural Language Processing}, pages 737--747, Copenhagen, Denmark. Association for Computational Linguistics.

\bibitem[{McNamee(2005)}]{McNamee}
Paul McNamee. 2005.
\newblock Language identification: A solved problem suitable for undergraduate instruction.
\newblock \emph{J. Comput. Sci. Coll.}, 20(3):94–101.

\bibitem[{Mende(2023)}]{+2023}
Jana-Katharina Mende, editor. 2023.
\newblock \href {https://doi.org/doi:10.1515/9783110778656} {\emph{Hidden Multilingualism in 19th-Century European Literature}}.
\newblock De Gruyter, Berlin, Boston.

\bibitem[{Stepanyan(2005)}]{stepanyan}
Hasmik Stepanyan. 2005.
\newblock \emph{(Armenian-Turkish-French) Bibliographie des livres et de la presse armeno-turque, 1727–1968}.
\newblock Istanbul: Turkuaz Yayınları.

\bibitem[{Werner(2012)}]{sarah}
Sarah Werner. 2012.
\newblock Where material book culture meets digital humanities.
\newblock \emph{Journal of Digital Humanities}, 1.

\end{thebibliography}

\end{document}